\documentclass[11pt, a4paper]{article}

\usepackage{pgfplotstable}
\usepackage{numprint}
\usepackage{array}
\usepackage{placeins}
\usepackage{mathtools}
\usepackage{setspace} 
\usepackage{dsfont}
\usepackage{amsfonts}
\usepackage{amsmath}
\usepackage[font=small]{caption}
\usepackage{subcaption}
\usepackage{paralist}
\usepackage{times}  
\usepackage{latexsym}
\usepackage{graphicx}
\usepackage[T1]{fontenc}
\usepackage{tikz}
\usepackage{url}
\usepackage{titlesec}
\usepackage{color}
\usepackage{lipsum,adjustbox}
\usepackage{bbm}
\usepackage{booktabs}

\newcolumntype{L}[1]{>{\raggedright\let\newline\\\arraybackslash\hspace{0pt}}m{#1}}
\newcolumntype{C}[1]{>{\centering\let\newline\\\arraybackslash\hspace{0pt}}m{#1}}
\newcolumntype{R}[1]{>{\raggedleft\let\newline\\\arraybackslash\hspace{0pt}}m{#1}}

\makeatletter
\newcommand{\@BIBLABEL}{\@emptybiblabel}
\newcommand{\@emptybiblabel}[1]{}
\usepackage[hidelinks]{hyperref}

\usepackage{acl2018}
\newcommand{\com}[1]{}

\newenvironment{myequation}{
	\vspace{-1em}
	\begin{equation}
}{
\end{equation}
\vspace{-1.2em}
}
\newenvironment{myequation*}{
	\vspace{-1em}
	\begin{equation*}
}{
\end{equation*}
\vspace{-1.2em}
}
\aclfinalcopy 

\begin{document}
	\title{Inherent Biases in Reference-based Evaluation for \\ Grammatical Error Correction and Text Simplification}
	\author{
		Leshem Choshen\textsuperscript{1} and Omri Abend\textsuperscript{2} \\
		\textsuperscript{1}School of Computer Science and Engineering,
		\textsuperscript{2} Department of Cognitive Sciences \\
		The Hebrew University of Jerusalem \\
		\texttt{leshem.choshen@mail.huji.ac.il, oabend@cs.huji.ac.il}\\
	}
	\maketitle
	
	\begin{abstract}
		The prevalent use of too few references for evaluating text-to-text
		generation is known to bias estimates of their quality (henceforth, {\it low
			coverage bias} or LCB). This paper shows that overcoming LCB in
		Grammatical Error Correction (GEC) evaluation cannot be attained by
		re-scaling or by increasing the number of references in any feasible
		range, contrary to previous suggestions. This is due to the long-tailed distribution of valid corrections for a sentence.
		Concretely, we show that LCB incentivizes GEC systems to avoid
		correcting even when they can generate a valid correction. 
		Consequently, existing systems obtain comparable or
		superior performance compared to humans, by making few but targeted 
		changes to the input.
		Similar effects on Text Simplification further support our claims.
	\end{abstract}
	
	\section{Introduction}
	
	Evaluation in monolingual translation \cite{xu2015problems,inderjeet2009summarization} and
	in particular in GEC
	\cite{tetreault2008native,madnani2011they,felice2015towards,bryant2015far,napoles2015ground}
	has gained notoriety for its difficulty, due in part to the heterogeneity and size of the space of 
	valid corrections \cite{chodorow2012problems,dreyer2012hyter}.
	Reference-based evaluation measures (RBM) are the common practice in GEC, including the standard
	$M^2$ \cite{dahlmeier2012better}, GLEU \cite{napoles2015ground} and I-measure \cite{felice2015towards}.
	
	The Low Coverage Bias (LCB) was previously discussed by \citet{bryant2015far}, 
	who showed that inter-annotator agreement in producing references is low, 
	and concluded that RBMs under-estimate the performance of GEC systems.
	To address this, they proposed a new measure, Ratio Scoring, which re-scales $M^2$ 
	by the inter-annotator agreement (i.e., the score of a human corrector), interpreted as an upper bound.
	
	We claim that the LCB has more far-reaching implications than previously discussed. 
	First, while we agree with \citet{bryant2015far} that a human correction should receive a perfect score, 
	we show that LCB does not merely scale system performance by a constant factor, but
	rather that some correction policies are less prone to be biased against. 
	Concretely, we show that by only correcting closed class errors, where few possible corrections are valid, systems can outperform humans. 
	Indeed, in Section \ref{sec:real_world} we show that some existing systems outperform humans on $M^2$ and GLEU, while only applying few changes to the source.
	
	We thus argue that the development of GEC systems against low coverage RBMs disincentivizes systems from making changes to the source 
	in cases where there are plentiful valid corrections (open class errors), as necessarily only some of them are covered by the reference set. 
	To support our claim we show that (1) existing GEC systems under-correct, often performing an order of magnitude less corrections than a human does (\S\ref{subsec:under-correction});
	(2) increasing the number of references alleviates under-correction (\S\ref{subsec:reranking});
	and (3) under-correction is more pronounced in error types that are more varied in their valid corrections (\S\ref{subsec:by_types}).
	
	A different approach for addressing LCB was taken by \cite{bryant2015far,sakaguchi2016reassessing},
	who propose to increase the number of references (henceforth, $M$).
	In Section \ref{sec:increase-reference} we estimate the distribution of corrections per sentence, 
	and find that increasing $M$ is unlikely to overcome LCB, due
	to the vast number of valid corrections for a sentence and their long-tailed distribution.
	Indeed, even short sentences have over 1000 valid corrections on average. 
	Empirically assessing the effect of increasing $M$ on the bias, we find diminishing returns using
	three standard GEC measures ($M^2$, accuracy and GLEU), underscoring the difficulty in this approach.
	
	Similar trends are found when conducting such experiments to Text Simplification (TS) (\S\ref{sec:simplification}).
	Specifically we show that (1) the distribution of valid simplifications for a given sentence is long-tailed; 
	(2) common measures for TS dramatically under-estimate performance; 
	(3) additional references alleviate this under-prediction.
	
	To recap, we find that the LCB hinders the reliability of RBMs for GEC,
	and incentivizes systems developed to optimize these measures not to correct. 
	LCB cannot be overcome by re-scaling or increasing $M$ in any feasible range. 
	\section{Coverage in RBMs}\label{sec:increase-reference}
	
	We begin by formulating a methodology for studying the distribution of valid 
	corrections for a sentence (\S\ref{subsec:corrections_distribution}), 
	and then turn to assessing the effect inadequate 
	coverage has on common RBMs (\S\ref{subsec:Assessment-values}). 
	Finally, we compare human and system scores by common RBMs (\S\ref{sec:real_world}).
	
	
	\paragraph{Notation.}
	We assume each ungrammatical sentence $x$ has a set of valid corrections $Correct_x$,
	and a discrete distribution $\mathcal{D}_x$ over them, where $P_{\mathcal{D}_x}(y)$
	for $y \in Correct_x$ is the probability a human annotator would correct $x$ as $y$.
	
	Let $X=x_{1}\ldots x_{N}$ be the evaluated set of source sentences and denote $\mathcal{D}_{i}\coloneqq \mathcal{D}_{x_i}$. Each $x_{i}$ is independently sampled from some 
	distribution $\mathcal{L}$ over input sentences, 
	and is paired with $M$ corrections $Y_i = \left\{y_{i}^{1},\ldots, y_{i}^{M}\right\}$,
	which are independently sampled from $\mathcal{D}_{i}$. Our analysis assumes a fixed number of references across sentences, 
	but generalizing to sentence-dependent $M$ is straightforward.
	The {\it coverage} of a reference set $Y_i$ of size $M$ for a sentence $x_i$ is defined as $P_{y \sim \mathcal{D}_i}(y \in Y_i)$.
	
	A system $C$ is a function from input sentences to proposed corrections (strings).
	An evaluation measure is a function $f\colon X \times Y \times C\to \mathbb{R}$. We use the term 
	``true measure'' to refer to a measure's output where the reference set includes all valid corrections, 
	i.e., $\forall i\colon Y_i=Correct_i$.
	
	\paragraph{Experimental Setup.}\label{par:experimental_setup}
	We conduct all experiments on the NUCLE test dataset \cite{dahlmeier2013building}.
	NUCLE is a parallel corpus of essays written by language learners and their corrected versions,
	containing 1414 essays and 50 test essays, each of about 500 words.
	
	We evaluate all participating systems in the CoNLL 2014 shared task,
	in addition to three of the best performing systems on this dataset,
	a hybrid system \cite{rozovskaya2016grammatical},
	a phrase-based MT system \cite{junczysdowmunt-grundkiewicz:2016:EMNLP2016} 
	and a neural network system \cite{xie2016neural}.
	Appendix  \ref{ap:abbr} lists system names and abbreviations.
	
	%
	\subsection{Estimating the Corrections Distribution}\label{subsec:corrections_distribution}
	%
	\paragraph{Data.}
	We turn to estimating the number of corrections per sentence, and their histogram.
	The experiments in the following section are run on a random sample of 52 short sentences from the NUCLE test data, i.e. with 15 words or less. Through the length restriction, we avoid introducing too many independent errors that may drastically increase the number of annotation variants (as every combination of corrections for these errors is possible), thus resulting in unreliable estimation for $\mathcal{D}_x$. 
	
	Proven effective in GEC and related tasks such as MT \cite{zaidan2011crowdsourcing,madnani2011they,post2012constructing}, 
	we use crowdsourcing to sample from $\mathcal{D}_x$ (see Appendix  \ref{ap:crowd}).
	Aiming to judge grammaticality rather than fluency, we instructed the workers to correct only when necessary, not for styling.
	We begin by estimating the histogram of $\mathcal{D}_x$ for each sentence, using the crowdsourced corrections.
	We use {\sc UnseenEst} \cite{zou2015quantifying}, a non-parametric algorithm to
	estimate a discrete distribution in which the individual values do not matter, only their probability. 
	{\sc UnseenEst} aims to minimize the ``earthmover distance'', between the estimated histogram and the histogram of the distribution. 
	Intuitively, if histograms are piles of dirt, {\sc UnseenEst} minimizes the amount of dirt moved times the distance it moved.
	{\sc UnseenEst} was originally developed and tested for estimating the histogram of
	variants a gene may have, including undiscovered ones, a setting similar to ours.
	Our manual tests of {\sc UnseenEst} with small artificially created datasets
	showed satisfactory results.\footnote{An implementation of \href{https://github.com/borgr/unseenest}{\sc UnseenEst}, the data we collected, the estimated distributions and efficient implementations of computations with \href{https://github.com/borgr/PoissonBinomial}{Poisson binomial distributions} can be found in \url{https://github.com/borgr/IBGEC}.}
	
	Our estimates show that most input sentences have a large number of
	infrequent corrections that account for much of the probability mass
	and a rather small number of frequent corrections.
	Table \ref{tab:corrections_dist} presents the mean number of different corrections with frequency at least $\gamma$ (for different $\gamma$s), and their total probability mass.
	For instance, 74.34 corrections account for 75\% of the probability mass, each occurring with frequency $\geq$ 0.1\%.
	
	\begin{table}[h!]
		\centering
		\small
		\singlespacing
		\begin{tabular}{c|c|c|c|c|}
			& \multicolumn{4}{c|}{Frequency Threshold ($\gamma$)}\\ 
			& \multicolumn{1}{c}{0} & \multicolumn{1}{c}{0.001} & \multicolumn{1}{c}{0.01} & \multicolumn{1}{c|}{0.1}
			\\
			\hline
			Variants & 1351.24 & 74.34 & 8.72 & 1.35
			\\
			Mass & 1 & 0.75 & 0.58 & 0.37\\
			\hline
		\end{tabular}
		\caption{\label{tab:corrections_dist}
			Estimating the distribution of corrections $\mathcal{D}_x$.
			The table presents the mean number of corrections per sentence with probability more than
			$\gamma$ (top row), as well as their total probability mass (bottom row).
		}
	\end{table}
	
	The high number of rare corrections raises the question of whether these can be regarded as noise.
	To test this we conducted another crowdsourcing experiment, where 3 annotators were asked to judge whether a correction produced in the first experiment, is indeed valid.
	We plot the validity of corrections against their frequencies, finding that frequency has little effect,
	where even the rarest corrections are judged valid 78\% of the time.
	Details in Appendix \ref{ap:validity_judgements}.
	
	\subsection{Under-estimation as a Function of $M$} \label{subsec:Assessment-values}
	
	After estimating the histogram of valid corrections for a sentence, we turn to estimating the resulting bias (LCB), for different $M$ values. 
	We study sentence-level accuracy, $F$-Score and GLEU.
	
	\paragraph{Sentence-level Accuracy.}
	Sentence-level accuracy is the percentage of corrections that
	exactly match one of the references.
	Accuracy is a basic, interpretable measure, used in GEC by, e.g., \newcite{rozovskaya2010annotating}.
	It is also closely related to the 0-1 loss function commonly used
	for training in GEC \cite{chodorow2012problems,rozovskaya2013joint}. 
	
	Formally, given test sentences $X=\{x_1,\ldots,x_N\}$,
	their references $Y_1,\ldots,Y_N$ and a system $C$,
	we define $C$'s accuracy to be
	
	\begin{small}
		\centering
		\begin{myequation}\label{eq:acc_def}
			Acc\left(C;X,Y\right) = \frac{1}{N} \sum_{i=1}^N \mathds{1}_{C(x_i) \in Y_i}.
		\end{myequation}
	\end{small}
	
	Note that $C$'s accuracy is, in fact, an estimate of $C$'s {\it true accuracy}, the probability to produce a valid correction for a sentence. Formally:
	
	\begin{small}
		\centering
		\begin{myequation}
			TrueAcc\left(C\right) = P_{x\sim{L}}\left(C\left(x\right)\in Correct_x\right).
		\end{myequation}
	\end{small}
	%
	
	The bias of $Acc\left(C;X,Y\right)$ for a sample of $N$ sentences, each paired with $M$ references
	is then
	
	\begin{small}
		\centering
		\begin{flalign}
		&TrueAcc\left(C\right) - \mathbb{E}_{X,Y}\left[Acc\left(C;X,Y\right)\right] = &\\
		&TrueAcc\left(C\right) - P\left(C\left(x\right) \in Y\right)  = &\\
		&P\left(C\left(x\right) \in Correct_x\right)  \cdot &\\
		&\label{eq:bias} \left(1 - P\left(C\left(x\right) \in Y \vert C\left(x\right) \in Correct_x\right) \right) &
		\end{flalign}
	\end{small}
	
	We observe that the bias, denoted $b_M$, is not affected by $N$, only by $M$.
	As $M$ grows, $Y$  better approximates $Correct_x$, and $b_M$ tends to 0.
	
	In order to abstract away from the idiosyncrasies of specific systems,
	we consider an idealized learner, which, when correct, produces a valid correction with the same
	distribution as a human annotator (i.e., according to $\mathcal{D}_x$).
	Formally, we assume that, if $C(x) \in Correct_x$ then $C(x) \sim \mathcal{D}_x$.
	Hence the bias $b_M$ (Eq. \ref{eq:bias}) can be re-written as
	
	\begin{small}
		\begin{myequation*}
			\centering
			P(C(x) \in Correct_x) \cdot (1 - P_{Y \sim \mathcal{D}_x^{M},y\sim \mathcal{D}_x}(y \in Y)).
		\end{myequation*}
	\end{small}
	
	We will henceforth assume that $C$ is perfect (i.e., its {\it true accuracy} is 1).
	Note that assuming any other value for $C$'s {\it true accuracy}
	would simply scale $b_M$ by that accuracy.
	Similarly, assuming only a fraction $p$ of the sentences require correction scales $b_M$ by $p$.
	%
	%
	
	We estimate $b_M$ empirically using its empirical mean on our experimental corpus:
	
	\begin{small}
		\begin{myequation*}
			\hat{b}_M = 1 - \frac{1}{N}\sum_{i=1}^N P_{Y \sim \mathcal{D}_i^M, y \sim \mathcal{D}_i}\left(y \in Y\right).
		\end{myequation*}
	\end{small}
	
	Using the {\sc UnseenEst} estimations of $\mathcal{D}_i$, we can compute $\hat{b}_M$ 
	for any size of $Y_i$ ($M$). 
	However, as this is highly computationally demanding, we estimate it using
	sampling. Specifically, for every $M = 1,...,20$ and $x_i$, we sample $Y_i$ 1000 times (with replacement), and estimate $P\left(y \in Y_i\right)$ as the covered probability mass $P_{\mathcal{D}_i}\{y: y \in Y_i\}$. Based on that we compute the accuracy distribution and expectation (see Appendix \ref{ap:poibin}).
	
	We repeated all our experiments where $Y_i$ is sampled without replacement,
	and find similar trends with a faster increase in accuracy reaching over $0.47$ with $M=10$.
	
	%
	%
	%
	%

	Figures \ref{fig:accuracy_vals} in supplementary \ref{ap:measures} presents the expected accuracy of a perfect
	system (i.e., 1-$\hat{b}_M$) for different  $M$s. 
	Results show that even for $M$ values which are much larger than the standard (e.g., $M=20$),
	expected accuracy is only around 0.5. As $M$ increases, the contribution of each additional correction 
	diminishes sharply (the slope is 0.004 for $M=20$).
	
	We also experiment with a more relaxed measure, {\it Exact Index Match}, which is only sensitive to the identity of the changed words and not to what they were changed to. 
	Formally, two corrections $c$ and $c'$ over a source sentence $x$ match if for their word alignments with the source (computed as above) $a:\{1,...,\left|x\right|\} \rightarrow \{1,...,\left|c\right|,Null\}$
	and $a':\{1,...,\left|x\right|\} \rightarrow \{1,...,\left|c'\right|,Null\}$, it holds that $c_{a\left(i\right)} \neq x_{i} \Leftrightarrow c'_{a'\left(i\right)} \neq x_{i}$, where $c_{Null}=c'_{Null}$.
	Results, while somewhat higher, are still only 0.54 with $M=10$. (Figure \ref{fig:accuracy_vals} in supplementray \ref{ap:measures})
	
	\paragraph{$F$-Score.}
	While accuracy is commonly used as a loss function for training GEC systems,
	$F_\alpha$-score is standard for evaluating system performance.
	The score is computed in terms of {\it edit} overlap between edits that constitute a correction and ones that constitute a reference, where edits are sub-string replacements to the source.
	We use the standard $M^2$ scorer \cite{dahlmeier2012better}, which defines edits optimistically, maximizing over all possible annotations that generate the correction 
	from the source. Since our crowdsourced corrections are not annotated for edits, we produce edits to the reference heuristically.
	
	The complexity of the measure prohibits an analytic approach \cite{yeh2000more}.
	We instead use bootstrapping to estimate the bias incurred by not being able to exhaustively enumerate the set of valid corrections.
	As with accuracy, in order to avoid confounding our results with system-specific biases,
	we assume the evaluated system is perfect and sample its corrections from the human distribution of corrections $\mathcal{D}_x$.
	
	Concretely, given a value for $M$ and for $N$, we uniformly sample from our experimental corpus source sentences $x_1,...,x_N$, and $M$ corrections for each $Y_1,...,Y_N$ (with replacement).
	Setting a realistic value for $N$ in our experiments is important for obtaining comparable results to those obtained on the NUCLE corpus (see \S\ref{sec:real_world}),
	as the expected value of $F$-score depends on $N$ and the number of sentences that do not need correction ($N_{cor}$).
	Following the statistics of NUCLE's test set, we set $N=1312$ and $N_{cor}=136$.
	
	Bootstrapping is carried out by the accelerated bootstrap procedure \cite{efron1987better}, with 1000 iterations.
	We also report confidence intervals ($p=.95$), computed using the same procedure.
	%
	
	Results (Figure \ref{fig:gleu_Ms} in supplementary \ref{ap:measures}) again show the insufficiency of commonly-used
	$M$ values for reliably estimating system performance.
	For instance, the $F_{0.5}$-score for our perfect system is only 0.42 with $M=2$.
	The saturation effect, observed for accuracy, is even more pronounced in this setting.
	
	\paragraph{GLEU.} We repeat the procedure using the mean GLEU sentence score (Figure \ref{fig:gleu_Ms} in supplementary \ref{ap:measures}), 
	which was shown to better correlate with human judgments than $M^2$ \cite{napoles-sakaguchi-tetreault:2016:EMNLP2016}.
	Results are about $2\%$ higher than $M^2$'s with a similar saturation effect.
	\citet{sakaguchi2016reassessing} observed a similar effect when evaluating against fluency-oriented references;
	this has led them to assume that saturation is due to covering most of the probability mass,
	which we now show is not the case.\footnote{
		We do not experiment with I-measure \cite{felice2015towards}, 
		as its run-time is prohibitively high for experimenting with bootstrapping
		that requires many applications of the measure \cite{choshen2018maege}, and as empirical validation 
		studies showed that it has a low correlation with human judgments \cite{sakaguchi2016reassessing}.}

	\subsection{Human and System Performance}\label{sec:real_world}
	
	The bootstrapping method for computing the significance of the $F$-score (\S\ref{subsec:Assessment-values}) 
	can also be used for assessing the significance of the differences in system performance reported in the literature.
	We compute confidence intervals of different systems on the NUCLE
	test data ($M=2$).
	
\begin{figure}
	\includegraphics[width=0.9\columnwidth]{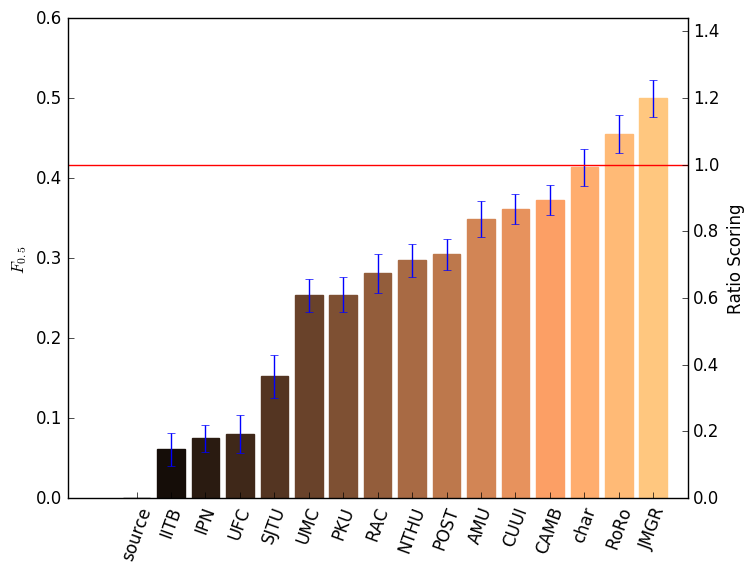}
	\caption{$F_{0.5}$ values with $M=2$ for different systems, including confidence interval ($p=.95$).
		The left-most column (``source'') presents the $F$-score of a system that doesn't make any
		changes to the source sentences. In red is human performance.
		See \S \ref{par:experimental_setup} for a legend of the systems.}\protect\label{fig:F_correctors}
\end{figure}

	Results (Figure \ref{fig:F_correctors}) present mixed trends: some
	differences between previously reported $F$-scores are indeed significant and some are not.
	For example, the best performing system is significantly better than all but the second one.
	
	Considering the $F$-score of the best-performing systems, and comparing 
	them to the $F$-score of a perfect system with $M=2$ (in accordance with systems' reported results),
	we find that their scores are comparable, where the systems RoRo and JMGR surpass a perfect system's $F$-score.
	Similar experiments with GLEU show that the two systems obtain comparable or superior performance to humans
	on this measure as well.
	
	%
	
	\subsection{Discussion}\label{subsec:mult_discussion}

	In this section we have established that (1) as systems can surpass human performance on RBMs, re-scaling cannot
	be used to overcome the LCB, and that (2) as the distribution of valid corrections is long-tailed, 
	the number of references needed for reliable RBMs is exceedingly high.
	Indeed, an average sentence has hundreds or more valid low-probability corrections, whose 
	total probability mass is substantial. 
	Our analysis with Exact Index Match suggests that similar effects are applicable to Grammatical Error Detection as well.
	The proposal of \newcite{sakaguchi2016reassessing}, to emphasize fluency over grammaticality in reference corrections, 
	only compounds this problem, as it results in a larger number of valid corrections.
	

	\section{Implications of the LCB}\label{sec:formal_conservatism}
	
	We discuss the adverse effects of LCB not only on the reliability of RBMs, but on the development
	of GEC systems.
	We argue that evaluation with inadequate reference coverage incentivizes systems to under-correct,
	and to mostly target errors that have few valid corrections (closed-class).
	We first show that low coverage can lead to under-correction (\S\ref{subsec:motivating_analysis}),
	then show that modern systems make far fewer corrections to the source, compared to humans (\S\ref{subsec:under-correction}).
	\S\ref{subsec:reranking} shows that increasing the number of references can alleviate this effect.
	\S\ref{subsec:by_types} shows that open-class errors are more likely to be under-corrected than closed-class ones.
	
	\subsection{Motivating Analysis}\label{subsec:motivating_analysis}
	
	For simplicity, we abstract away from the details of the learning model and assume 
	that systems attempt to maximize an objective function, 
	over some training or development data. 
	We assume maximization is achieved by iterating over the samples, as with the Perceptron or SGD.
	
	Assume the system is faced with a phrase it predicts to be ungrammatical. 
	Assume $p_{detect}$ is the probability this prediction is correct, and
	$p_{correct}$ is the probability it is able to predict
	a valid correction for this phrase (including correctly identifying it as erroneous).
	Finally, assume evaluation is
	against $M$ references with coverage $p_{coverage}$
	(the probability that
	a valid correction will be found among $M$ randomly sampled references).
	
	We will now assume that the system may either choose to correct with the correction it finds 
	the most likely or not at all. If it chooses not to correct, its probability of being rewarded 
	(i.e., its output is in the reference set) is $(1-p_{detect})$. Otherwise, its probability
	of being rewarded is $p_{correct} \cdot p_{coverage}$.
	A system is disincentivized from altering the phrase in cases where:
	
	\vspace{.2cm}
	\begin{small}
		\begin{myequation}
			\label{eq:reward}
			p_{correct} \cdot p_{coverage} < 1-p_{detect} 
		\end{myequation}
	\end{small}

	We expect Condition (\ref{eq:reward}) to frequently hold in cases that
	require non-trivial changes, which are characterized both by low $p_{coverage}$ (as non-trivial
	changes are often open-class), and by lower system performance.
	
	Precision-oriented measures (e.g., $F_{0.5}$) penalize invalidly correcting more
	harshly than not correcting an ungrammatical sentence.
	In these cases, Condition (\ref{eq:reward}) should be written as
	
	\begin{small}
		\begin{myequation*}
			p_{correct} \cdot p_{coverage} - \left(1-p_{correct} \cdot p_{coverage}\right) \alpha < 1-p_{detect} 
		\end{myequation*}
	\end{small}
	
	\noindent
	where $\alpha$ is the ratio between the penalty for introducing a wrong correction and the reward for a valid correction. The condition is even more likely to hold with such measures.

	\subsection{Under-correction in GEC Systems}\label{subsec:under-correction}
	\begin{table}
		\centering
		\small
		\singlespacing
		\begin{tabular}{l|p{4cm}}
			Corrector & Sentence \\
			\hline
			Source & This is especially to people who are overseas. \\
			\hline 
			CHAR, UMC, JMGR & This is especially \textbf{for} people who are overseas. \\ 
			IPN & This is especially to \textbf{peoples} who are overseas. \\ 
			CUUI &  This is especially to \textbf{the} people who are overseas. \\ 
			\hline
			NUCLE$_A$ & This is especially \textbf{true for} people who are overseas.\\
			NUCLE$_B$ & This is especially \textbf{relevant} to people who are overseas.
		\end{tabular}
		\caption{\label{tab:nucle_example} Example for a sentence and proposed corrections by different systems (top part) and by the two NUCLE annotators (bottom part). Systems not mentioned in the table retain the source. No system produces a new word as needed. The two references differ in their corrections.} 
	\end{table} 
	
	In this section we compare the prevalence of changes made to the source by the systems,
	to their prevalence in the NUCLE references. 
	To strengthen our claim, we exclude all non-alphanumeric characters, 
	both within tokens or as separate tokens. See Table \ref{tab:nucle_example} 
	for an example.
	
	We consider three types of divergences between the source and the reference.
	First, we measure the extent to which \emph{words} were changed: altered, deleted or added.
	To do so, we compute word alignment between the source and the reference, casting it
	as a weighted bipartite matching problem. Edge weights are assigned to 
	be the token edit distances.\footnote{Aligning words in GEC is much simpler than in MT, as most of the words are unchanged, deleted fully, added, or changed slightly.}
	Following word alignment, we define {\sc WordChange}
	as the number of aligned words and unaligned words changed.
	Second, we quantify word \emph{order} differences using
	Spearman's $\rho$ between the order of the words in the source sentence
	and the order of their corresponding-aligned words in the correction.
	$\rho=0$ where the word order is uncorrelated, and $\rho=1$ where the orders exactly match. We report the average $\rho$ over all source sentence pairs. 
	Third, we report how many source sentences were split and how many concatenated by the reference and by the systems.
	One annotator was arbitrarily selected for the figures.

	\paragraph{Results.} \hspace{-.5cm}
	Results (Figures in supplementary \ref{fig:over-conservatism}) show that humans make considerably more changes than systems according to all measures of under-correction, 
	both in terms of the number of sentences modified and the number of modifications within them. Differences are often an order of magnitude large.
	For example, 36 reference sentences include 6 word changes, where the maximal number of sentences with 6 word changes by any system is 5.
	We find similar trends on the references of the TreeBank of Learner English \cite{yannakoudakis2011new}.
	
	\subsection{Higher $M$ Alleviates Under-correction}\label{subsec:reranking}
	
	This section reports an experiment for determining whether increasing
	the number of references in training indeed reduces under-correction. There is no 
	corpus available with multiple references which is large enough for re-training a system. 
	Instead, we simulate such a setting with an oracle reranking approach, and test whether the 
	availability of increasingly more training references reduces a system's under-correction.
	
	Concretely, given a set of sentences, each paired with $M$ references, a measure and a 
	system's $k$-best list, we define an oracle re-ranker that selects for each sentence the highest scoring correction.
	As a test case, we use the RoRo system with $k=100$, and apply it to the 
	largest available language learner corpus which is paired with a substantial amount of GEC references,
	namely the NUCLE test corpus. We use the standard $F$-score as the evaluation measure,
	examining the under-correction of the oracle re-ranker for different $M$ values, averaging over the 1312 samples of 
	$M$ references from the available set of ten references provided by \citet{bryant2015far}.
	
	As the argument is not trivial, we turn to explaining why decreased under-correction with an increase in $M$ indicates
	that tuning against a small set of references (low coverage) yields under-correction. 
	Assume an input sentence with some sub-string $e$. 
	There are three cases: (1) $e$ is an error, (2) $e$ is valid but there are valid references that alter it, (3) $e$ is uniquely valid. In case (3) oracle re-ranking has no effect and can be ignored.
	The corrections in the $k$-best list can then be partitioned to those that keep $e$ as it is; those that invalidly alter $e$; and those that validly alter $e$.

	\begin{table}[t]
		\centering
		\small
		\singlespacing
		\begin{tabular}{c|C{1cm}|cc}
			& $L_{val}$ empty & \multicolumn{2}{c}{$L_{val}$ not empty} \\
			&            & $e$ valid & $e$ error \\ \hline
			\multicolumn{1}{c|}{Small $M$} & 0 & $P_Y(\overline{e}, L_{val})$ & $P_Y\left(L_{val}\right)$    \\
			\multicolumn{1}{c|}{Large $M$} & 0 & 0              & 1                  \\ \hline
			\multicolumn{1}{c|}{Correction Rate}            & =  & $\downarrow$        & $\uparrow$
		\end{tabular}
		\caption{\label{ta:oracle_expected_results}
			The expected effect of oracle re-ranking on under-correction.
			Values represent the probability of altering a sub-string of the input $e$, which is a proxy to the expected correction rate. $L_{val}$ is the valid 
			alterations in the $k$-best list. $P_Y\left(L_{val}\right)$ is the probability that a valid correction from the list is also in the reference set $Y$,
			$P_Y(\overline{e}, L_{val})$ is the probability that, in addition, the reference that keeps $e$ is not in $Y$.
			When $M$ increases, the expected correction rate is expected to increase only if $e$ is an error and a valid correction of it
			is found in the $k$-best list.
		}

	\end{table}
	
	Table \ref{ta:oracle_expected_results} presents the probability that $e$ will be altered in the different cases.
	Analysis shows that under-correction is likely to decrease with $M$ only
	in the case where $e$ is an error and the $k$-best list contains a valid correction of it.
	Whenever the reference allows both keeping $e$ and altering $e$, the re-ranker selects keeping $e$. 
	
\begin{figure}
		\includegraphics[width=\columnwidth]{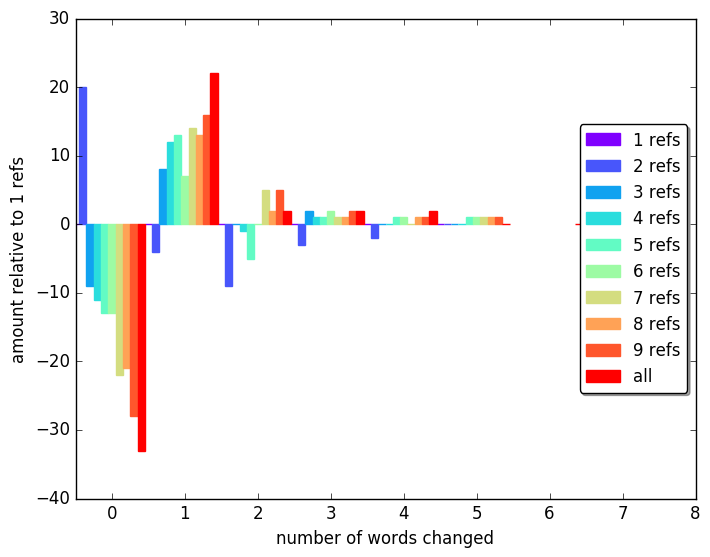}
		\caption{The amount of sentences (y-axis) with a given number of words changed (x-axis) following oracle reranking with different $M$ values (column colors), 
			where the amount for $M=1$ is subtracted from them.
			All references are randomly sampled except the ``all'' column that contains all ten references.
			In conclusion, tuning against additional references indeed reduces under-correction.
			\label{fig:reranking_word_change}
		}
	\end{figure}

	
	Indeed, our experimental results show that word changes increase with $M$ (Figure \ref{fig:reranking_word_change}),
	indicating that low coverage may play a role in the observed tendency of GEC systems to under-correct.
	No significant difference is found for word order.

	%
	%

	\subsection{Under-correction by Error Types}\label{subsec:by_types}
	
	In this section we study the prevalence of under-correction according to edit types, finding that open-class types of errors
	(such as replacing a word with another word) are more starkly under-corrected, than closed-class errors.
	Evaluating with low coverage RBMs does not incentivize systems to address open-class errors (in fact, it disincentivizes them to). Therefore, even if LCB is not the cause for this trend, current evaluation procedures may perpetuate it.
	
	We use the data of \citet{bryant-felice-briscoe:2017:Long}, which automatically assigned types to each edit 
	in the output of all CoNLL 2014 systems on the NUCLE test set.
	As a measure of under-correction tendency, we take the ratio between the mean number of corrections produced by the systems and by the references.
	We note that this analysis does not consider whether the predicted correction is valid or not, but only how many of the errors of each type the systems attempted to correct. 
	
	We find that all edit types are under-predicted on average, but that the least under-predicted ones are mostly closed-class types. 
	Concretely, the top quarter of error types consists of orthographical errors, 
	plurality inflection of nouns, 
	adjective inflections to superlative or comparative forms and determiner selection. The bottom quarter includes the categories 
	verb selection, noun selection, particle/preposition selection, pronoun selection, and the type {\sc OTHER}, which is a residual category.
	The only exception to this regularity is the closed-class punctuation selection type, which is found in the lower quarter. See Appendix \ref{ap:types}.
	
	This trend cannot be explained by assuming that common error types are targeted more.
	Indeed, error type frequency is slightly negatively correlated with the under-correction ratio ($\rho$=-0.29 p-value=0.16).
	A more probable account of this effect is the disincentive of GEC systems to correct open-class error types, for which even valid
	corrections are unlikely to be rewarded.
	
	\section{Similar Effects on Simplification}\label{sec:simplification}
	
	We now turn to replicating our experiments on Text Simplification (TS). 
	From a formal point of view, evaluation of the tasks is similar:
	the output is obtained by making zero or more edits to the source. RBMs are the standard for TS evaluation,
	much like they are in GEC.
	
	Our experiments on TS demonstrate that similar trends recur in this setting as well. 
	The tendency of TS systems to under-predict changes to the source 
	has already been observed by previous work \cite{alvamanchego-EtAl:2017:I17-1}, 
	showing that TS systems under-predict word additions, deletions,
	substitutions, and sequence shifts \cite{zhang-lapata:2017:EMNLP2017},
	and have low edit distance from the source \cite{narayan-gardent:2016:INLG}.
	Our experiments show that LCB may account for this under-prediction. Concretely, we show that
	(1) the distribution of valid references for a given sentence is long-tailed; 
	(2) common evaluation measures suffer from LCB, taking SARI \cite{Xu-EtAl:2016:TACL} 
	as an example RBM (similar trends are obtained with Accuracy); 
	(3) under-prediction is alleviated with $M$ in oracle re-ranking experiments.
	
	We crowd-sourced 2500 reference simplifications for 47 sentences, using the corpus and the annotation protocol of 
	\newcite{Xu-EtAl:2016:TACL}, and applying {\sc UnseenEst} to estimate $\mathcal{D}_x$ (Appendix  \ref{ap:crowd}).
	Table \ref{tab:simplifications_dist} shows that the expected number of references is even greater in this setting. 
	
	Assessing the effect of $M$ on SARI, we find that SARI diverges from Accuracy and $F$-score
	in that its multi-reference version is not a maximum over the single-reference scores, but some combination of them.
	This can potentially increase coverage, but it also leads to an unintuitive situation: an output 
	identical to a reference does not receive a perfect score, but rather the score depends on how similar the output is to the other
	references. A more in-depth analysis of SARI's handling of multiple references is found in Appendix \ref{ap:sari-assum}.
	In order to neutralize this effect of SARI, we also report results with MAX-SARI, which coincides with SARI on $M=1$, 
	and is defined as the maximum single-reference SARI score for $M>1$.
	
	\com{
		\begin{figure}
			\includegraphics[width=0.9\columnwidth]{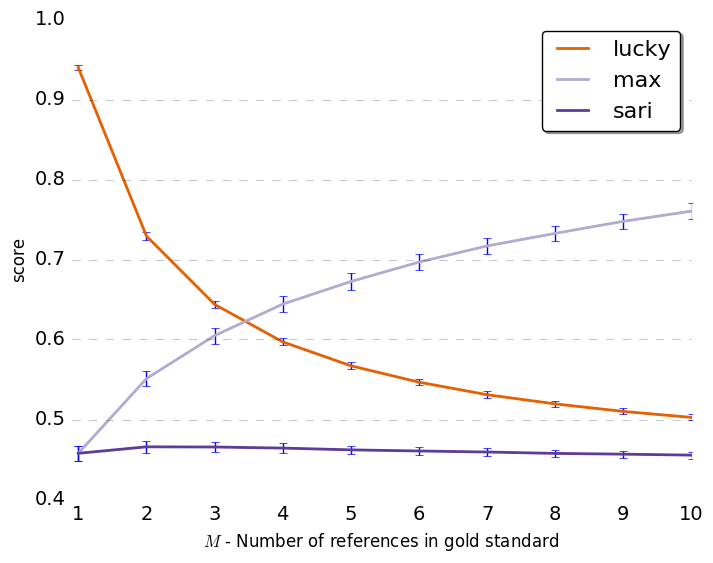}
			\caption{
				SARI and MAX-SARI values for a perfect system and SARI values for a ``lucky perfect'' system (y-axis) as a function of the number of references $M$ (x-axis).
				Each data point is paired with a confidence interval ($p=.95$).\label{fig:SARI_Ms}}
		\end{figure}
	}
	
	Figure \ref{fig:SARI_Ms} in supplementary \ref{ap:measures} presents the coverage of SARI and MAX-SARI of a perfect TS system that selects a random correction
	from the estimated distribution of corrections using the same bootstrapping protocol as in \S\ref{subsec:corrections_distribution}.
	We also include the SARI score of a ``lucky perfect'' system, that randomly selects one of 
	the given references (the MAX-SARI score for such a system is 1). 
	Results show that SARI has a coverage of about $0.45$, and that this score is largely independent of $M$.
	The score of predicting one of the available references drops with the number of references, indicating that SARI scores may not be comparable
	across different $M$ values. 
	
	We therefore restrict oracle re-ranking experiments to MAX-SARI, conducting re-ranking experiments on $k$-best
	lists in two settings: Moses \cite{koehn2007koehn} 
	with $k=100$, and a neural model \cite{nisioi2017exploring} with $k=12$. 
	Our results indeed show that under-prediction is alleviated with $M$ in both settings. 
	For example, the least under-predicting model (the neural one) did not change 50 sentences with $M=1$, but only 29 weren't changed 
	with $M=8$. See Appendix \ref{ap:simp-rerank}.
	
	
	
	
	\begin{table}[t]
		\centering
		\small
		\singlespacing
		\begin{tabular}{c|c|c|c|c|}
			& \multicolumn{4}{c|}{Frequency Threshold ($\gamma$)}\\ 
			& \multicolumn{1}{c}{0} & \multicolumn{1}{c}{0.001} & \multicolumn{1}{c}{0.01} & \multicolumn{1}{c|}{0.1}
			\\
			\hline
			Variants & 2636.29 & 111.19 & 4.68 & 0.13
			\\
			Mass & 1 & 0.42 & 0.14 & 0.02\\
			\hline
		\end{tabular}
		\caption{\label{tab:simplifications_dist}
			Estimating the distribution of simplifications $\mathcal{D}_x$.
			The table presents the mean number of simplifications per sentence with probability more than
			$\gamma$ (top row), as well as their total probability mass (bottom row).
		}
	\end{table}
	\section{Conclusion}\label{sec:conclusion}
	
	
	
	We argue that using low-coverage reference sets has adverse effects on the reliability
	of reference-based evaluation, with GEC and TS as a test case, and consequently on the incentives offered to systems.
	We further argue that these effects cannot be overcome by re-scaling or increasing the number of references in a feasible way. 
	The paper makes two methodological contributions to the monolingual translation evaluation literature:
	(1) a methodology for evaluating evaluation measures by the scores they assign a perfect system, using a bootstrapping procedure;
	(2) a methodology for assessing the distribution of valid monolingual translations.
	Our findings demonstrate how these tools can help characterize the biases of existing systems and evaluation measures.
	We believe our findings and methodologies can be useful for similar tasks such as 
	style conversion and automatic post-editing of raw MT outputs.
	
	We note that the LCB further jeopardizes the reliability of common validation experiments for RBMs,
	that assess the correlation between human and measure rankings of system outputs \cite{grundkiewicz2015human}.
	Indeed, if outputs all similarly under-correct, correlation studies will not be affected by whether an RBM is sensitive to under-correction.
	Therefore, the tendency of RBMs to reward under-correction
	cannot be detected by such correlation experiments \cite[cf.][]{choshen2018maege}.
	
	Our results underscore the importance of developing alternative evaluation measures that transcend $n$-gram overlap, 
	and use deeper analysis tools, e.g., by comparing
	the semantics of the reference and the source to the output \cite[cf.][]{lo2011meant}.
	\newcite{napoles-sakaguchi-tetreault:2016:EMNLP2016}
	have made progress towards this goal in proposing a reference-less grammaticality measure,
	using Grammatical Error Detection tools, as did \newcite{asano2017reference}, who added a fluency measure to the grammaticality.
	In a recent project \cite{choshen2018usim}, we proposed a complementary measure that 
	measures the semantic faithfulness of the output to the source, in order to form a combined semantic measure that bypasses the pitfalls of low coverage.

	\section*{Acknowledgments}
	
	This work was supported by the Israel Science Foundation (grant No. 929/17),
	and by the HUJI Cyber Security Research Center in conjunction with the Israel
	National Cyber Bureau in the Prime Minister's Office.
	We thank Nathan Schneider, Courtney Napoles and Joel Tetreault for helpful feedback.
	
	\bibliographystyle{acl_natbib}
	\bibliography{propose}
	
	\appendix
	
	\onecolumn
		\FloatBarrier
	\section{Measure scores by references}\label{ap:measures}
\begin{figure}[]
		\centering
		\includegraphics[width=0.5\textwidth]{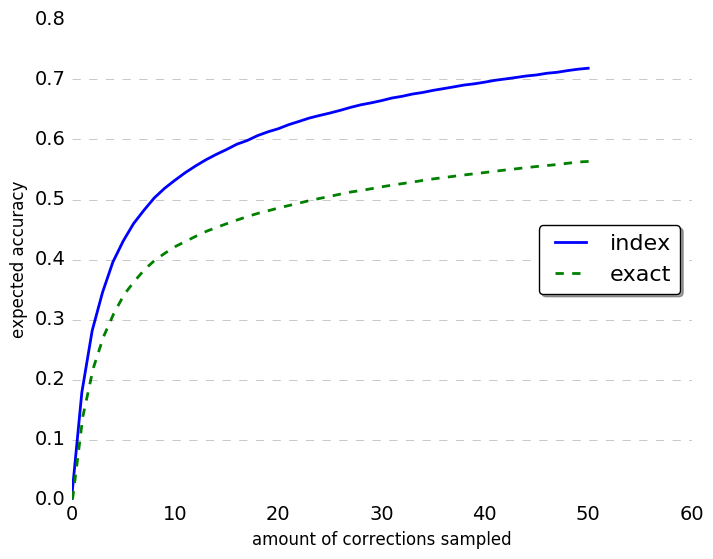}
		\caption{Accuracy and Exact Index Match.
		} \protect\label{fig:accuracy_vals}
				
	\end{figure}

\begin{figure}[]
		\centering
		\includegraphics[width=0.5\textwidth]{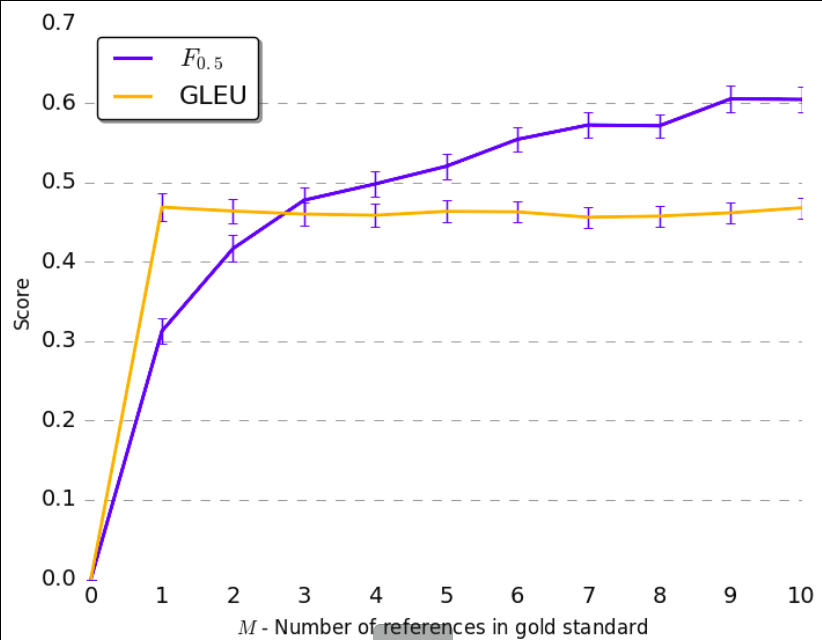}
		\caption{GLEU and F}\protect\label{fig:gleu_Ms}
	\end{figure}

	The score obtained by perfect systems according to GEC accuracy (\protect\ref{fig:accuracy_vals}), GEC F-score and GLEU (\protect\ref{fig:gleu_Ms}).
	Figure \ref{fig:SARI_Ms} reports TS experimental results, namely the score of a perfect and lucky perfect system using SARI,
	and a perfect system using MAX-SARI. The y-axis corresponds to the measure values, and the x-axis to the number of references $M$.
	For bootstrapping experiments points are paired with a confidence interval ($p=.95$).
	
	\begin{figure}[]
		\centering
		\includegraphics[width=0.5\textwidth]{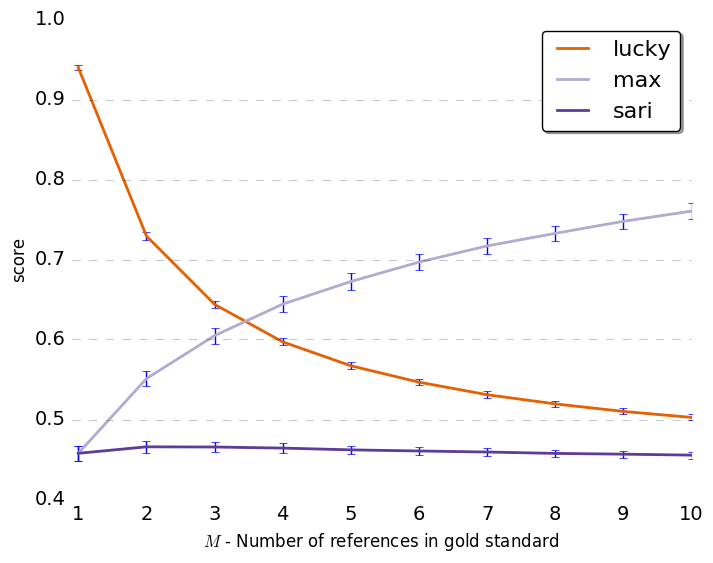}
		\caption{
			(lucky) perfect SARI and MAX-SARI }\protect\label{fig:SARI_Ms}
	\end{figure}

	\FloatBarrier
		\section{Amount of corrections by systems}
		\protect\label{fig:over-conservatism}
		The prevalence of changes in system outputs and in the NUCLE reference.
		The top figure presents the number of sentences (heat) for each amount of word changes
		(x-axis; measured by {\sc WordChange}) done by the outputs and the reference (y-axis).
		The middle figure presents the percentage of sentence pairs (y-axis) where the
		Spearman $\rho$ values do not exceed a certain threshold (x-axis).
		The bottom figure presents the counts of source sentences (y-axis) concatenated (right bars) or split (left bars) by the references (striped column) and the outputs (coloured columns).
		See Appendix \protect\ref{ap:abbr} for a legend of the systems.
		Under all measures, the gold standard references make substantially more changes to the source sentences than any of the systems, in some cases an order of magnitude more.
		\begin{figure}[]
			\includegraphics[width = .5\textwidth]{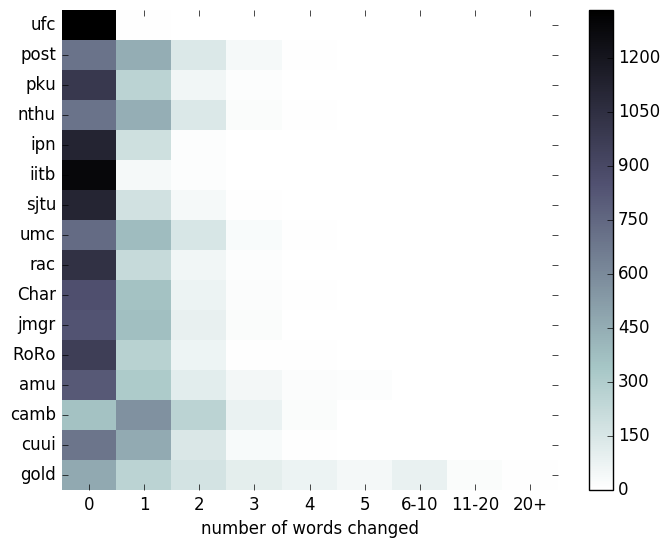}
		\end{figure}
		
		\begin{figure}[]
			\includegraphics[width = .5\textwidth]{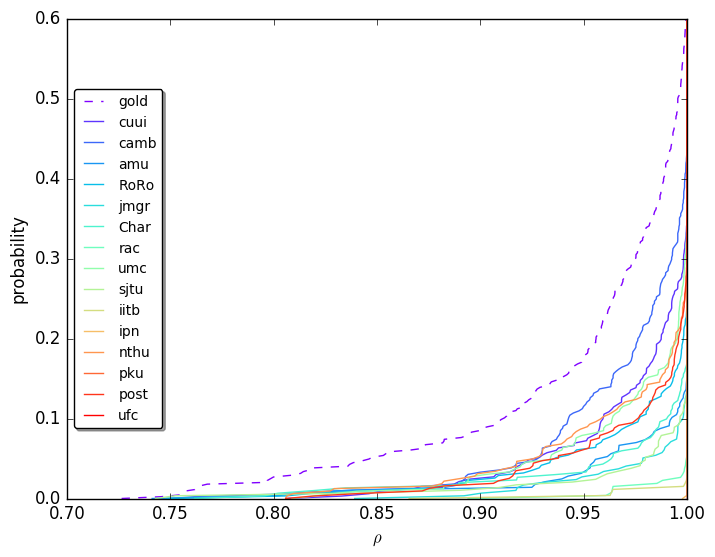}
		\end{figure}
		
		\begin{figure}[]
			\includegraphics[width = .5\textwidth]{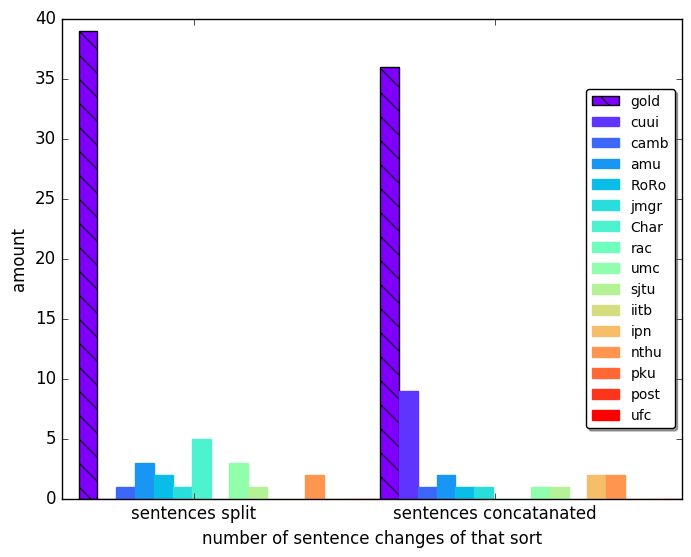}
		\end{figure}
	
		\FloatBarrier
	\section{Systems tested}\label{ap:abbr}
	Adam Mickiewicz University (AMU),
	University of Cambridge (CAMB), Columbia University and the University of Illinois at Urbana-Champaign (CUUI),
	Indian Institute of Technology, Bombay (IITB), Instituto Politecnico Nacional (IPN),
	National Tsing Hua University (NTHU), Peking University (PKU), Pohang University of Science and Technology (POST),
	Research Institute for Artificial Intelligence, Romanian Academy (RAC), Shanghai Jiao Tong University (SJTU),
	University of Franche Comt\'{e} (UFC), University of Macau (UMC), \newcite[RoRo]{rozovskaya2016grammatical}, \newcite[JMGR]{junczysdowmunt-grundkiewicz:2016:EMNLP2016} \newcite[Char]{xie2016neural}.
	
	\section*{\centering\Large Supplementary Material for ``Inherent Biases in Reference-based Evaluation for
		Grammatical Error Correction''}

	\section{Collected references}\label{ap:crowd}
	\subsection{Grammatical Error Correction}
	\begin{table}[]
		\centering
		\begin{tabular}{ll}
			\cline{1-1}
			\multicolumn{1}{|l|}{origin} & Other relatives may have the same possibilities to have such kind of disease .        \\ \hline
			1                            & It is possible other relatives may have the same kind of disease .                    \\
			2                            & It is possible that other relatives may have the same kind of disease .               \\
			3                            & It's also possible for other relatives to have the same kind of disease.              \\
			4                            & Other relatives may also be predisposed to the same kind of diseases.                 \\
			5                            & Other relatives may be at risk to have the same disease.                              \\
			6                            & Other relatives may have be prone to having such diseases.                            \\
			7                            & Other relatives may have similar possibilities to have the same disease.              \\
			8                            & Other relatives may have the possibility of having the same kind of disease.          \\
			9                            & Other relatives may have the possibility to have the same such disease .              \\
			10                           & Other relatives may have the same chance of contracting the same kind of disease.     \\
			11                           & Other relatives may have the same chance of having that kind of disease.              \\
			12                           & Other relatives may have the same chance of suffering such diseases.                  \\
			13                           & Other relatives may have the same chances of having the same kind of disease.         \\
			14                           & Other relatives may have the same chance to have the same disease.                    \\
			15                           & Other relatives may have the same likelihood of having such a disease.                \\
			16                           & Other relatives may have the same possibilities of having such a disease .            \\
			17                           & Other relatives may have the same possibilities of having such a disease .            \\
			18                           & Other relatives may have the same possibilities of having such a disease.             \\
			19                           & Other relatives may have the same possibilities of having that kind of disease.       \\
			20                           & Other relatives may have the same possibilities of having the same kind of disease.   \\
			21                           & Other relatives may have the same possibilities to develop such a disease.            \\
			22                           & Other relatives may have the same possibilities to have such a disease .              \\
			23                           & Other relatives may have the same possibilities to have such a kind of disease.       \\
			24                           & Other relatives may have the same possibilities to have such a kind of disease.       \\
			25                           & Other relatives may have the same possibilities to have such kinds of diseases.       \\
			26                           & Other relatives may have the same possibilities to have the same kind of disease .    \\
			27                           & Other relatives may have the same possibilities to have this kind of disease .        \\
			28                           & Other relatives may have the same possibilities to have this kind of disease.         \\
			29                           & Other relatives may have the same possibility of developing the same kind of disease. \\
			30                           & Other relatives may have the same possibility of having said disease.                 \\
			31                           & Other relatives may have the same possibility of having the disease.                  \\
			32                           & Other relatives may have the same possibility of having this kind of disease .        \\
			33                           & Other relatives may have the same possibility to have such a kind of disease .        \\
			34                           & Other relatives may have the same possibility to have such kind of a disease .        \\
			35                           & Other relatives may have the same possibility to have such kind of disease .          \\
			36                           & Other relatives may have the same possibility to have such kinds of disease .         \\
			37                           & Other relatives may have the same possibility to have such kinds of diseases.         \\
			38                           & Other relatives may have the same possibility to have the same kind of disease.       \\
			39                           & Other relatives may have the same probability of having this disease .                \\
			40                           & Other relatives may have the same probability to have the same kind of disease.       \\
			41                           & Other relatives may have the same risk for developing such a disease.                 \\
			42                           & Other relatives may have the same risk of having such a disease.                      \\
			43                           & Other relatives may have the same risk of having the disease.                         \\
			44                           & Other relatives may have the same strong possibility to have such kinds of disease.   \\
			45                           & Other relatives may possibly also have the same disease .                             \\
			46                           & Other relatives may possibly have the same risk of having that kind of disease.       \\
			47                           & Other relatives may possibly have the same such kind of disease .                     \\
			48                           & Relatives may also be more prone to similar diseases.                                 \\
			49                           & Relatives may have the same possibilities to have such kind of disease .              \\
			50                           & Those who are related may have the same chances of acquiring certain diseases.       
		\end{tabular}
		\caption{An example of one of the learner language sentences with the highest number of different corrections. The origin sentence is on top.}
	\end{table}
	52 sentences with a maximum length of 15 were collected from the NUCLE test corpus \cite{dahlmeier2013building}. 
	For each of the 52 source sentences, 
	we elicited 50 corrections from Amazon Mechanical Turk workers. Workers were native speakers (located in the US) having at least 1000 approved HITs and 98\% acceptance rate.
	Sentences with less than 6 words were discarded, as they were mostly a result of sentence segmentation errors.
	4 sentences required no correction according to almost half the workers and were hence discarded.
	Common methods based on agreement such as Fleiss's kappa or test questions as the problem we deal with is exactly the low agreement of valid corrections.
	4 workers were rejected due to suspicious answers. such were workers that most or all of their work changed nothing except non-alphanumeric characters and were the only ones to keep several source sentences unchanged. 
	
	\begin{figure}[h!]
		\includegraphics[width=0.9\columnwidth]{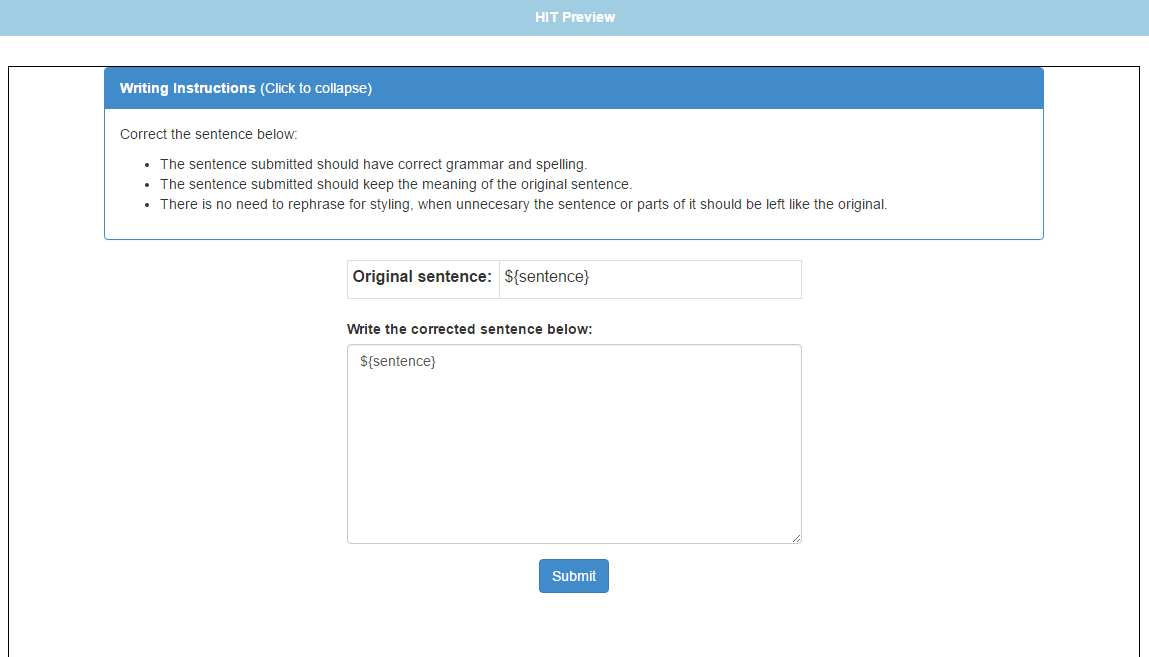}
		\caption{Template for a grammatical error correction annotation task} 
	\end{figure}
	\FloatBarrier
	
	\section{Validity}\label{ap:validity_judgements}
	\begin{figure}[h!]
		\vspace{-.3cm}
		\includegraphics[width=0.9\columnwidth]{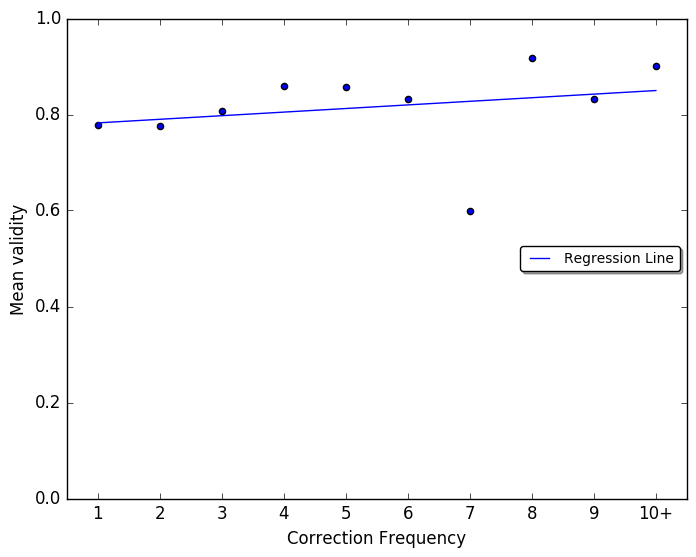}
		\caption{The mean frequency ($y$-axis) in which a correction that was produced
			a given number of times ($x$-axis), was judged to be valid.
			\label{fig:validity_judgements}}
		\vspace{-0.3cm}
	\end{figure}
	\FloatBarrier
	
	\section{Accuracy - Poisson Binomial Distribution}\label{ap:poibin}
	The analytic tools we have developed support the computation of the entire distribution of the accuracy, and not only its expected values. From the Equation 
	
	\begin{myequation}
		Acc\left(C;X,Y\right) = \frac{1}{N} \sum_{i=1}^N \mathds{1}_{C(x_i) \in Y_i}.
	\end{myequation}
	
	\noindent
	we see that Accuracy has a Poisson Binominal distribution (i.e., it is a sum of independent Bernoulli variables with different success probabilities), whose success probabilities are $P_{y,Y \sim \mathcal{D}_i}(y \in Y)$, which can be computed, as before, using our estimate for $\mathcal{D}_i$. Estimating the density function allows for the straightforward definition of significance tests for the measure, and can be performed efficiently \cite{hong2013computing}. An implementation of this and other methods for efficiently computing and approximating Poisson Binomial Distributions and the estimated density functions can be found in \url{https://github.com/borgr/PoissonBinomial}.
	
	\section{Type conservatism and prevalence}\label{ap:types}
	
	\begin{table}[]
		\npdecimalsign{.}
		\nprounddigits{3}
		\centering
		\resizebox{0.3\textwidth}{!}{%
			\begin{tabular}{@{}ln{5}{2}@{}}
				\toprule
				\multicolumn{1}{c}{TYPE}       & \multicolumn{1}{c}{CHANGE}       \\ \midrule
				CONTR      & 0.9705882353 \\
				NOUN:NUM   & 0.9545454545 \\
				ORTH       & 0.8925       \\
				ADJ:FORM   & 0.8101265823 \\
				NOUN:INFL  & 0.7125       \\
				DET        & 0.6724801812 \\
				VERB:SVA   & 0.6434231379 \\
				MORPH      & 0.6150712831 \\
				VERB:FORM  & 0.5384615385 \\
				SPELL      & 0.5373737374 \\
				VERB:INFL  & 0.5          \\
				ADJ        & 0.4596273292 \\
				CONJ       & 0.4303797468 \\
				ADV        & 0.416091954  \\
				PREP       & 0.3343130051 \\
				WO         & 0.3148148148 \\
				NOUN:POSS  & 0.2904564315 \\
				VERB:TENSE & 0.2887673956 \\
				NOUN       & 0.2823240589 \\
				PART       & 0.268        \\
				PRON       & 0.2391033624 \\
				PUNCT      & 0.2012539185 \\
				OTHER      & 0.1736497941 \\
				VERB       & 0.1509009009 \\ \bottomrule
			\end{tabular}%
		}
		\caption{Number of mean corrections of systems divided by the number of corrections by references on NUCLE dataset\cite{dahlmeier2012better}. Data is based on \citet{bryant-felice-briscoe:2017:Long}.}
	\end{table}
	\FloatBarrier
	
	\section{SARI with Multiple references}\label{ap:sari-assum}
	Most RBMs define their multi-reference score to be the maximum score the
	output attains against any single one reference.
	SARI takes a different approach and combines multiple references to yield its score, 
	possibly in order to compensate for the necessarily limited coverage of the available
	references. This yields non-monotonic behavior of SARI with respect to increasing
	the number of references, which makes the biases incurred by low coverage less predictable,
	but not less significant.
	For instance, a set of diverse set of references would span a large space of possible
	combinations, making SARI more permissive. A set of references of the same size, which only differ little
	from one another would yield a more conservative SARI score.
	
	SARI is defined as the average of 3 scores, based on how well the system output kept the words it should keep,
	deleted the words it should delete, and added the words it should add (all with respect to the reference).
	Each of these scores behaves differently when increasing the number of references.
	For a perfect recall for addition can only be obtained if all additions suggested by any reference are added. For a perfect precision for addition all additions should be found in at least one reference, acting similar to max operation. For a perfect precision on keeping all the references should keep all agree on keeping everything the system kept, and as we showed reference don't tend to agree on it. For a perfect recall the system should everything that at least one of the references chose to keep. Note, that the two terms together mean that for a perfect keep score a necessary condition is that all \textbf{references} should agree on what to keep. A perfect precision of deletion acts similarly to the one over keeping, but being precision oriented SARI ignores the deletion recall. Thus, a perfect deletion would be one to delete only words, and at least one, that all references agree on.

	\section{Reference Crowdsourcing for Simplification}\label{ap:simp-rerank}
	In order to perform our simplification experiments, 
	we collected additional references for sentences from the corpus presented in \newcite{Xu-EtAl:2016:TACL}, using Amazon
	Mechanical Turk with similar annotation guidelines to those used by Xu et al. 
	Overall we collect 50 references for 45, and 100 and 150 references for two more sentences.
	The latter two have 70 and 126 references that occur only once and none reoccurring more than 6 times, 
	supporting the claim that the size of the space of valid references for TS is huge.
	Standard worker rejection techniques were used, but no worker had to be rejected.
	
	\begin{figure}[h!]
		\includegraphics[width=0.9\columnwidth]{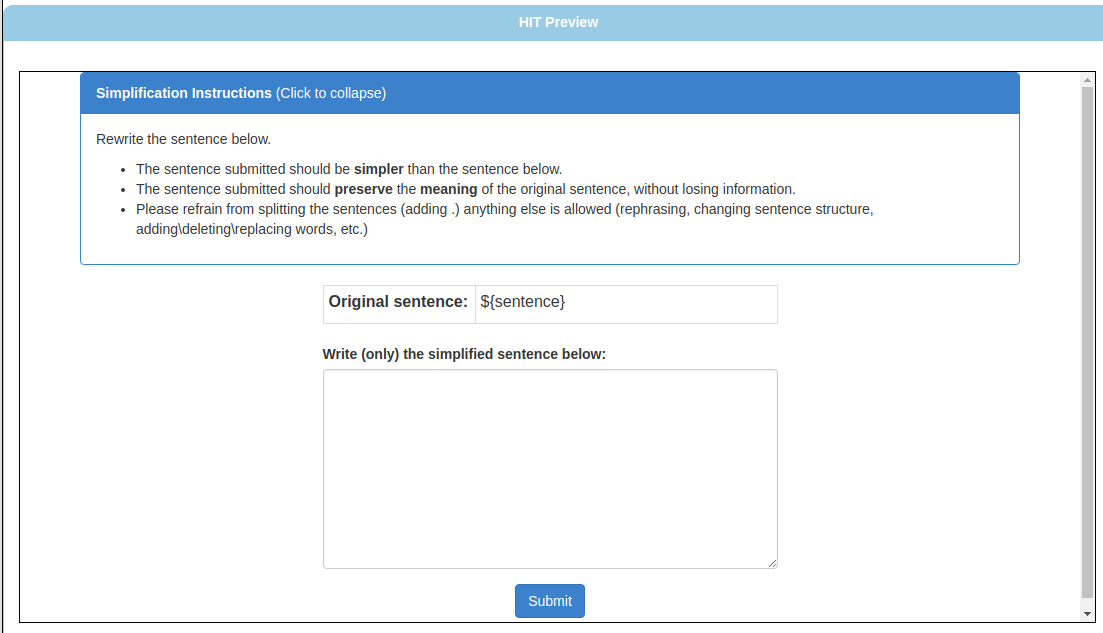}
		\caption{Template for a simplification annotation task} 
	\end{figure}
	\FloatBarrier
	\subsection{Simplification reranking}
	\begin{figure}[h!]
		\includegraphics[width=0.9\columnwidth]{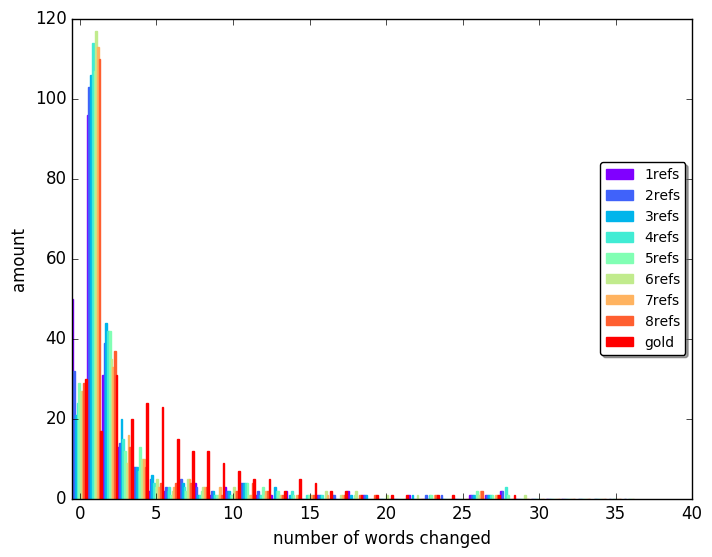}
		\caption{Reranking results with \newcite{nisioi2017exploring} and MAX-SARI}
	\end{figure}
	\begin{figure}[h!]
		\includegraphics[width=0.9\columnwidth]{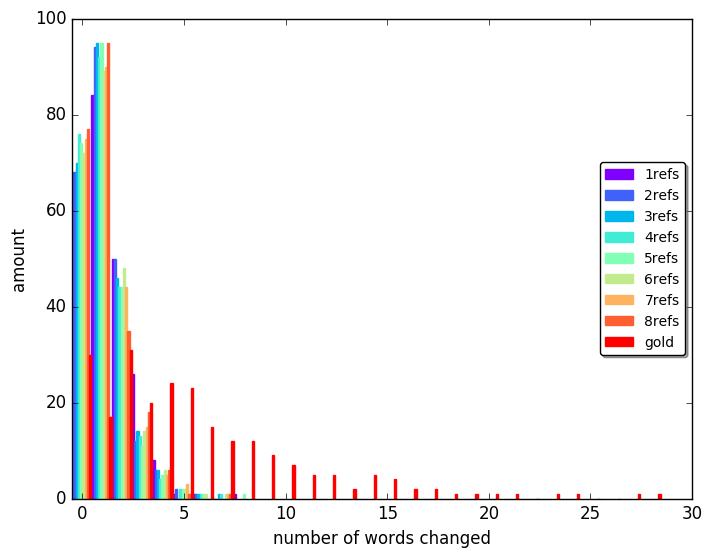}
		\caption{Reranking results with Moses and SARI}
		
	\end{figure}
	\begin{figure}[h!]
		\includegraphics[width=0.9\columnwidth]{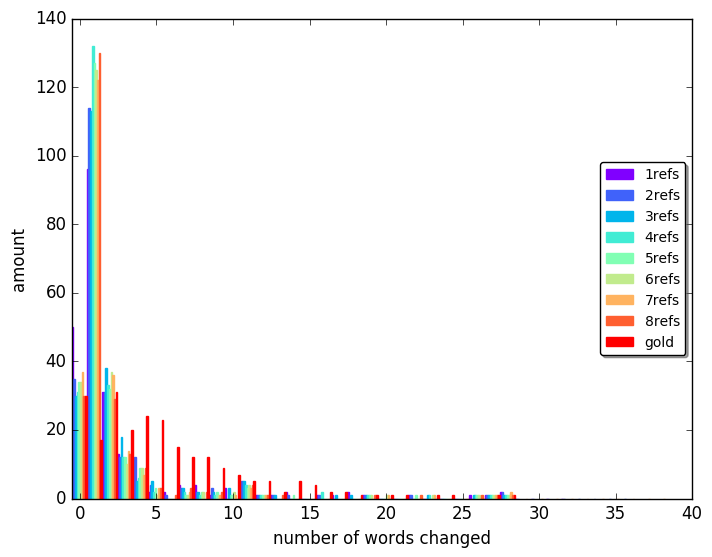}
		\caption{Reranking results with \newcite{nisioi2017exploring} and SARI}
		
	\end{figure}
	\begin{figure}[h!]
		\includegraphics[width=0.9\columnwidth]{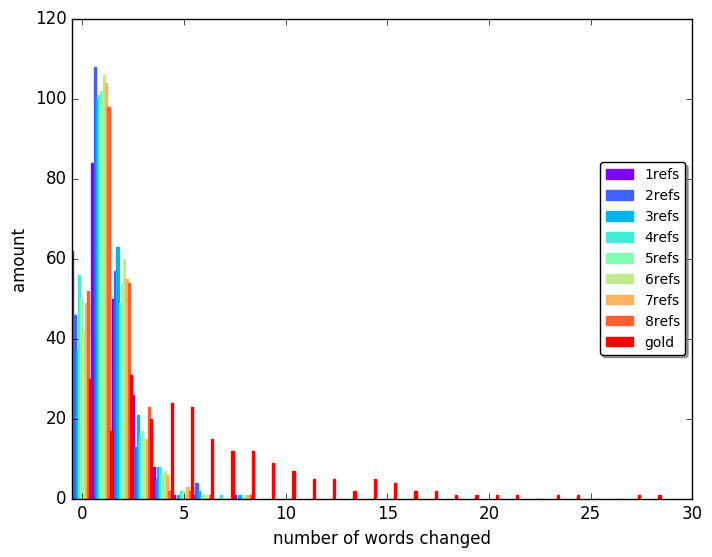}
		\caption{Reranking results with Moses and MAX-SARI}
		
	\end{figure}
	\FloatBarrier
	\com{
		\section{Annotated paragraphs}
		\begin{table}[hb]
			\centering
			\begin{tabular}{lll}
				Annotator-id & NUCLE-id & type      \\
				1         & 2  & corrected \\
				2         & 2  & corrected \\
				1         & 2  & learner   \\
				2         & 2  & learner   \\
				1         & 3  & corrected \\
				2         & 3  & corrected \\
				1         & 3  & learner   \\
				2         & 3  & learner   \\
				1         & 5  & corrected \\
				2         & 5  & corrected \\
				1         & 5  & learner   \\
				2         & 5  & learner   \\
				1         & 6  & learner   \\
				2         & 6  & learner   \\
				2         & 7  & corrected \\
				2         & 7  & learner   \\
				1         & 8  & corrected \\
				1         & 8  & learner   \\
				1         & 10 & corrected \\
				1         & 10 & learner  
			\end{tabular}
			\caption{The list of paragraphs annotated, showing which annotator annotated it, which type of language is used in it and the corresponding id in the NUCLE corpus. Note that parallel paragraphs have the same id.\label{tab:annotated-paragraphs}}
		\end{table}
	}
	\FloatBarrier

\end{document}


\appendix

\title{}

\onecolumn

\section{Systems tested}\label{ap:abbr}
Adam Mickiewicz University (AMU),
University of Cambridge (CAMB), Columbia University and the University of Illinois at Urbana-Champaign (CUUI),
Indian Institute of Technology, Bombay (IITB), Instituto Politecnico Nacional (IPN),
National Tsing Hua University (NTHU), Peking University (PKU), Pohang University of Science and Technology (POST),
Research Institute for Artificial Intelligence, Romanian Academy (RAC), Shanghai Jiao Tong University (SJTU),
University of Franche Comt\'{e} (UFC), University of Macau (UMC), \newcite[RoRo]{rozovskaya2016grammatical}, \newcite[JMGR]{junczysdowmunt-grundkiewicz:2016:EMNLP2016} \newcite[Char]{xie2016neural}.

\section*{\centering\Large Supplementary Material for ``Inherent Biases in Reference-based Evaluation for
Grammatical Error Correction''}

\section{Collected references}\label{ap:crowd}
\subsection{Grammatical Error Correction}
\begin{table}[]
	\centering
	\begin{tabular}{ll}
		\cline{1-1}
		\multicolumn{1}{|l|}{origin} & Other relatives may have the same possibilities to have such kind of disease .        \\ \hline
		1                            & It is possible other relatives may have the same kind of disease .                    \\
		2                            & It is possible that other relatives may have the same kind of disease .               \\
		3                            & It's also possible for other relatives to have the same kind of disease.              \\
		4                            & Other relatives may also be predisposed to the same kind of diseases.                 \\
		5                            & Other relatives may be at risk to have the same disease.                              \\
		6                            & Other relatives may have be prone to having such diseases.                            \\
		7                            & Other relatives may have similar possibilities to have the same disease.              \\
		8                            & Other relatives may have the possibility of having the same kind of disease.          \\
		9                            & Other relatives may have the possibility to have the same such disease .              \\
		10                           & Other relatives may have the same chance of contracting the same kind of disease.     \\
		11                           & Other relatives may have the same chance of having that kind of disease.              \\
		12                           & Other relatives may have the same chance of suffering such diseases.                  \\
		13                           & Other relatives may have the same chances of having the same kind of disease.         \\
		14                           & Other relatives may have the same chance to have the same disease.                    \\
		15                           & Other relatives may have the same likelihood of having such a disease.                \\
		16                           & Other relatives may have the same possibilities of having such a disease .            \\
		17                           & Other relatives may have the same possibilities of having such a disease .            \\
		18                           & Other relatives may have the same possibilities of having such a disease.             \\
		19                           & Other relatives may have the same possibilities of having that kind of disease.       \\
		20                           & Other relatives may have the same possibilities of having the same kind of disease.   \\
		21                           & Other relatives may have the same possibilities to develop such a disease.            \\
		22                           & Other relatives may have the same possibilities to have such a disease .              \\
		23                           & Other relatives may have the same possibilities to have such a kind of disease.       \\
		24                           & Other relatives may have the same possibilities to have such a kind of disease.       \\
		25                           & Other relatives may have the same possibilities to have such kinds of diseases.       \\
		26                           & Other relatives may have the same possibilities to have the same kind of disease .    \\
		27                           & Other relatives may have the same possibilities to have this kind of disease .        \\
		28                           & Other relatives may have the same possibilities to have this kind of disease.         \\
		29                           & Other relatives may have the same possibility of developing the same kind of disease. \\
		30                           & Other relatives may have the same possibility of having said disease.                 \\
		31                           & Other relatives may have the same possibility of having the disease.                  \\
		32                           & Other relatives may have the same possibility of having this kind of disease .        \\
		33                           & Other relatives may have the same possibility to have such a kind of disease .        \\
		34                           & Other relatives may have the same possibility to have such kind of a disease .        \\
		35                           & Other relatives may have the same possibility to have such kind of disease .          \\
		36                           & Other relatives may have the same possibility to have such kinds of disease .         \\
		37                           & Other relatives may have the same possibility to have such kinds of diseases.         \\
		38                           & Other relatives may have the same possibility to have the same kind of disease.       \\
		39                           & Other relatives may have the same probability of having this disease .                \\
		40                           & Other relatives may have the same probability to have the same kind of disease.       \\
		41                           & Other relatives may have the same risk for developing such a disease.                 \\
		42                           & Other relatives may have the same risk of having such a disease.                      \\
		43                           & Other relatives may have the same risk of having the disease.                         \\
		44                           & Other relatives may have the same strong possibility to have such kinds of disease.   \\
		45                           & Other relatives may possibly also have the same disease .                             \\
		46                           & Other relatives may possibly have the same risk of having that kind of disease.       \\
		47                           & Other relatives may possibly have the same such kind of disease .                     \\
		48                           & Relatives may also be more prone to similar diseases.                                 \\
		49                           & Relatives may have the same possibilities to have such kind of disease .              \\
		50                           & Those who are related may have the same chances of acquiring certain diseases.       
	\end{tabular}
	\caption{An example of one of the learner language sentences with the highest number of different corrections. The origin sentence is on top.}
\end{table}
52 sentences with a maximum length of 15 were collected from the NUCLE test corpus \cite{dahlmeier2013building}. 
For each of the 52 source sentences, 
we elicited 50 corrections from Amazon Mechanical Turk workers. Workers were native speakers (located in the US) having at least 1000 approved HITs and 98\% acceptance rate.
Sentences with less than 6 words were discarded, as they were mostly a result of sentence segmentation errors.
4 sentences required no correction according to almost half the workers and were hence discarded.
Common methods based on agreement such as Fleiss's kappa or test questions as the problem we deal with is exactly the low agreement of valid corrections.
4 workers were rejected due to suspicious answers. such were workers that most or all of their work changed nothing except non-alphanumeric characters and were the only ones to keep several source sentences unchanged. 

\begin{figure}[h!]
	\includegraphics[width=0.9\columnwidth]{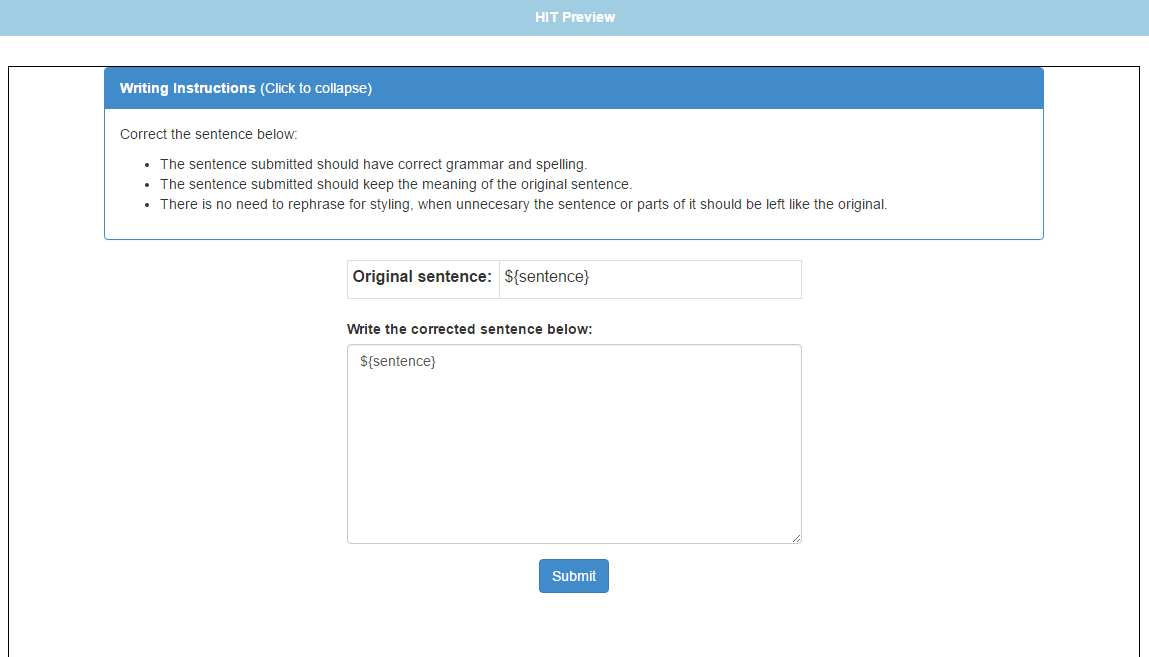}
	\caption{Template for a grammatical error correction annotation task} 
\end{figure}
\FloatBarrier

\section{Validity}\label{ap:validity_judgements}
\begin{figure}[h!]
	\vspace{-.3cm}
	\includegraphics[width=0.9\columnwidth]{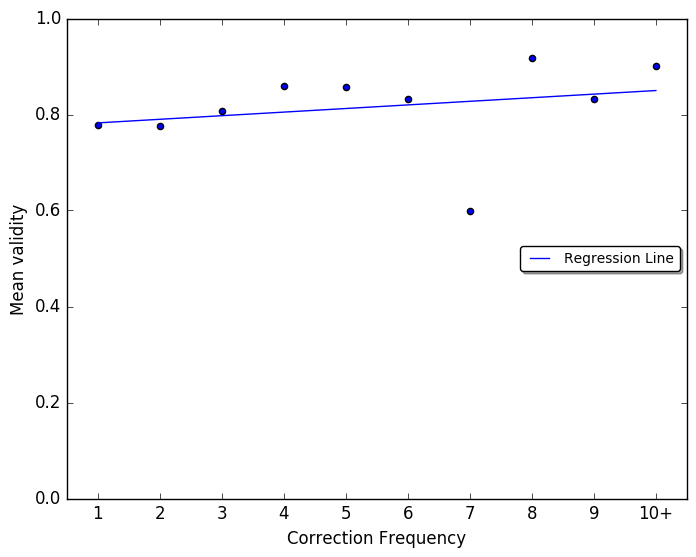}
	\caption{The mean frequency ($y$-axis) in which a correction that was produced
		a given number of times ($x$-axis), was judged to be valid.
		\label{fig:validity_judgements}}
	\vspace{-0.3cm}
\end{figure}
\FloatBarrier

\section{Accuracy - Poisson Binomial Distribution}\label{ap:poibin}
The analytic tools we have developed support the computation of the entire distribution of the accuracy, and not only its expected values. From the Equation 
	
	\begin{myequation}
		Acc\left(C;X,Y\right) = \frac{1}{N} \sum_{i=1}^N \mathds{1}_{C(x_i) \in Y_i}.
	\end{myequation}
	
\noindent
 we see that Accuracy has a Poisson Binominal distribution (i.e., it is a sum of independent Bernoulli variables with different success probabilities), whose success probabilities are $P_{y,Y \sim \mathcal{D}_i}(y \in Y)$, which can be computed, as before, using our estimate for $\mathcal{D}_i$. Estimating the density function allows for the straightforward definition of significance tests for the measure, and can be performed efficiently \cite{hong2013computing}. An implementation of this and other methods for efficiently computing and approximating Poisson Binomial Distributions and the estimated density functions can be found in \url{https://github.com/borgr/PoissonBinomial}.

\section{Type conservatism and prevalence}\label{ap:types}

\begin{table}[]
	\npdecimalsign{.}
	\nprounddigits{3}
	\centering
	\resizebox{0.3\textwidth}{!}{%
		\begin{tabular}{@{}ln{5}{2}@{}}
			\toprule
			\multicolumn{1}{c}{TYPE}       & \multicolumn{1}{c}{CHANGE}       \\ \midrule
			CONTR      & 0.9705882353 \\
			NOUN:NUM   & 0.9545454545 \\
			ORTH       & 0.8925       \\
			ADJ:FORM   & 0.8101265823 \\
			NOUN:INFL  & 0.7125       \\
			DET        & 0.6724801812 \\
			VERB:SVA   & 0.6434231379 \\
			MORPH      & 0.6150712831 \\
			VERB:FORM  & 0.5384615385 \\
			SPELL      & 0.5373737374 \\
			VERB:INFL  & 0.5          \\
			ADJ        & 0.4596273292 \\
			CONJ       & 0.4303797468 \\
			ADV        & 0.416091954  \\
			PREP       & 0.3343130051 \\
			WO         & 0.3148148148 \\
			NOUN:POSS  & 0.2904564315 \\
			VERB:TENSE & 0.2887673956 \\
			NOUN       & 0.2823240589 \\
			PART       & 0.268        \\
			PRON       & 0.2391033624 \\
			PUNCT      & 0.2012539185 \\
			OTHER      & 0.1736497941 \\
			VERB       & 0.1509009009 \\ \bottomrule
		\end{tabular}%
	}
	\caption{Number of mean corrections of systems divided by the number of corrections by references on NUCLE dataset\cite{dahlmeier2012better}. Data is based on \citet{bryant-felice-briscoe:2017:Long}.}
\end{table}
\FloatBarrier

\section{SARI with Multiple references}\label{ap:sari-assum}
Most RBMs define their multi-reference score to be the maximum score the
output attains against any single one reference.
SARI takes a different approach and combines multiple references to yield its score, 
possibly in order to compensate for the necessarily limited coverage of the available
references. This yields non-monotonic behavior of SARI with respect to increasing
the number of references, which makes the biases incurred by low coverage less predictable,
but not less significant.
For instance, a set of diverse set of references would span a large space of possible
combinations, making SARI more permissive. A set of references of the same size, which only differ little
from one another would yield a more conservative SARI score.

SARI is defined as the average of 3 scores, based on how well the system output kept the words it should keep,
deleted the words it should delete, and added the words it should add (all with respect to the reference).
Each of these scores behaves differently when increasing the number of references.
For a perfect recall for addition can only be obtained if all additions suggested by any reference are added. For a perfect precision for addition all additions should be found in at least one reference, acting similar to max operation. For a perfect precision on keeping all the references should keep all agree on keeping everything the system kept, and as we showed reference don't tend to agree on it. For a perfect recall the system should everything that at least one of the references chose to keep. Note, that the two terms together mean that for a perfect keep score a necessary condition is that all \textbf{references} should agree on what to keep. A perfect precision of deletion acts similarly to the one over keeping, but being precision oriented SARI ignores the deletion recall. Thus, a perfect deletion would be one to delete only words, and at least one, that all references agree on.

\section{Reference Crowdsourcing for Simplification}\label{ap:simp-rerank}
In order to perform our simplification experiments, 
we collected additional references for sentences from the corpus presented in \newcite{Xu-EtAl:2016:TACL}, using Amazon
Mechanical Turk with similar annotation guidelines to those used by Xu et al. 
Overall we collect 50 references for 45, and 100 and 150 references for two more sentences.
The latter two have 70 and 126 references that occur only once and none reoccurring more than 6 times, 
supporting the claim that the size of the space of valid references for TS is huge.
Standard worker rejection techniques were used, but no worker had to be rejected.

\begin{figure}[h!]
	\includegraphics[width=0.9\columnwidth]{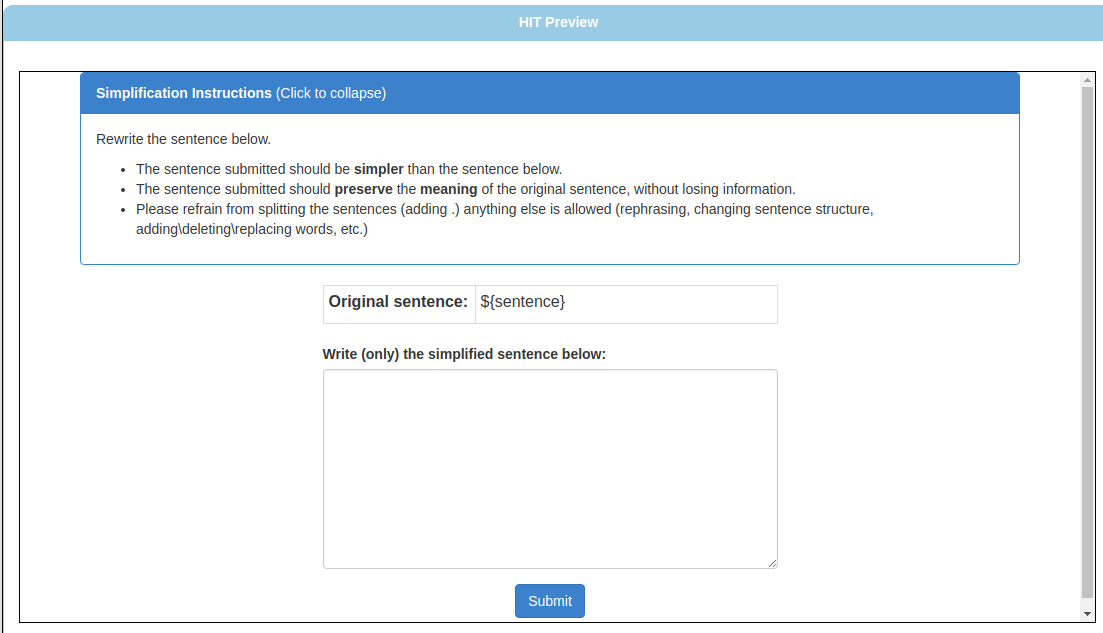}
	\caption{Template for a simplification annotation task} 
\end{figure}
\FloatBarrier
\subsection{Simplification reranking}
\begin{figure}[h!]
	\includegraphics[width=0.9\columnwidth]{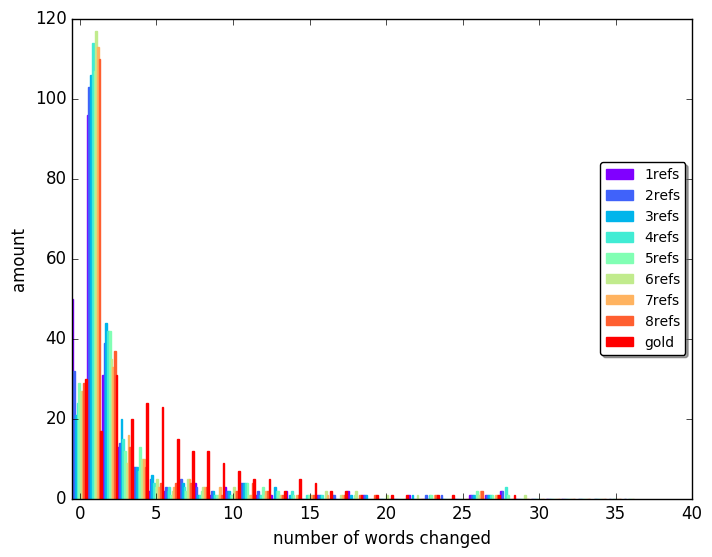}
	\caption{Reranking results with \newcite{nisioi2017exploring} and MAX-SARI}
\end{figure}
\begin{figure}[h!]
	\includegraphics[width=0.9\columnwidth]{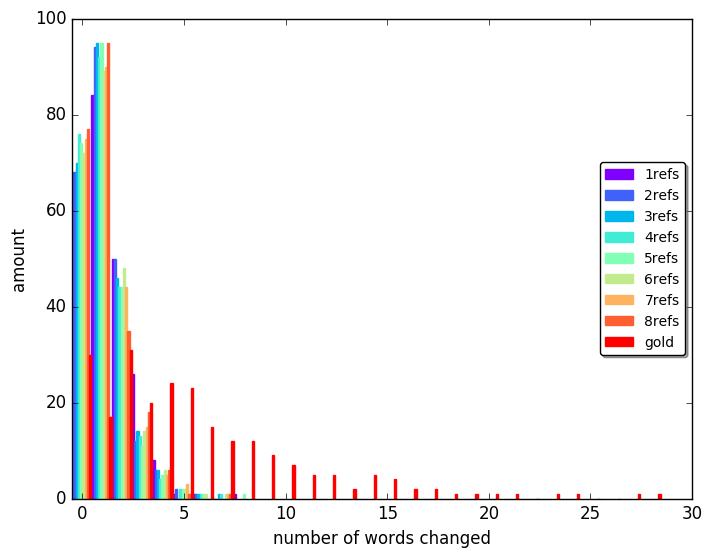}
	\caption{Reranking results with Moses and SARI}
	
\end{figure}
\begin{figure}[h!]
	\includegraphics[width=0.9\columnwidth]{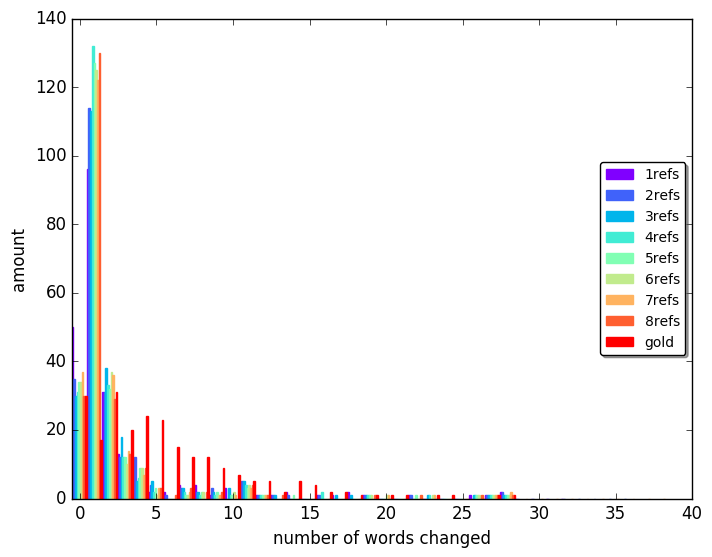}
	\caption{Reranking results with \newcite{nisioi2017exploring} and SARI}
	
\end{figure}
\begin{figure}[h!]
	\includegraphics[width=0.9\columnwidth]{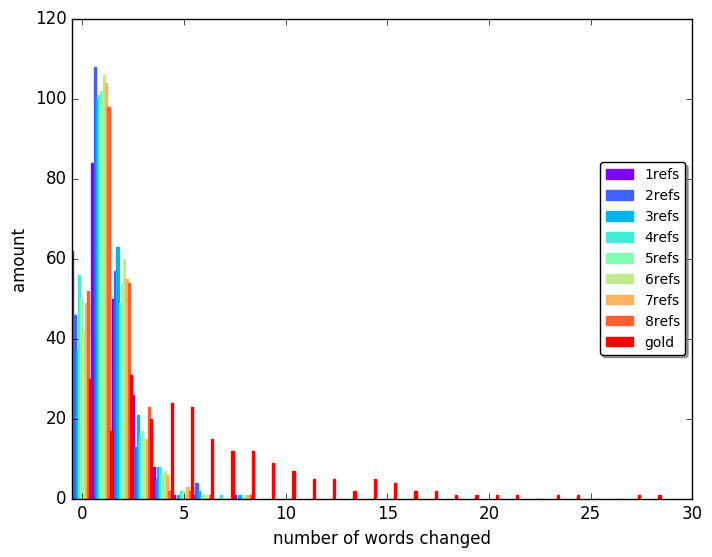}
	\caption{Reranking results with Moses and MAX-SARI}
	
\end{figure}
\FloatBarrier
\com{
	\section{Annotated paragraphs}
	\begin{table}[hb]
		\centering
		\begin{tabular}{lll}
			Annotator-id & NUCLE-id & type      \\
			1         & 2  & corrected \\
			2         & 2  & corrected \\
			1         & 2  & learner   \\
			2         & 2  & learner   \\
			1         & 3  & corrected \\
			2         & 3  & corrected \\
			1         & 3  & learner   \\
			2         & 3  & learner   \\
			1         & 5  & corrected \\
			2         & 5  & corrected \\
			1         & 5  & learner   \\
			2         & 5  & learner   \\
			1         & 6  & learner   \\
			2         & 6  & learner   \\
			2         & 7  & corrected \\
			2         & 7  & learner   \\
			1         & 8  & corrected \\
			1         & 8  & learner   \\
			1         & 10 & corrected \\
			1         & 10 & learner  
		\end{tabular}
		\caption{The list of paragraphs annotated, showing which annotator annotated it, which type of language is used in it and the corresponding id in the NUCLE corpus. Note that parallel paragraphs have the same id.\label{tab:annotated-paragraphs}}
	\end{table}
}
\FloatBarrier

\bibliographystyle{acl_natbib}
\bibliography{propose}